\documentclass[11pt]{article}

\usepackage{times}
\usepackage{latexsym}
\usepackage{graphicx}

\usepackage{url}
\usepackage{hyperref}
\usepackage{multicol}
\usepackage{caption}
\usepackage{amsmath}

\usepackage{microtype}
\usepackage[numbers]{natbib}

\title{Automated Sentiment Classification and Topic Discovery in Large-Scale Social Media Streams}

\date{}

\author{Yiwen Lu, Siheng Xiong, Zhaowei Li\\
Georgia Institute of Technology\\
\texttt{\{ylu647, sxiong45, zli899\}@gatech.edu}
}

\begin{document}

\maketitle

\begin{abstract}
We present a framework for large-scale sentiment and topic analysis of Twitter discourse. Our pipeline begins with targeted data collection using conflict-specific keywords, followed by automated sentiment labeling via multiple pre-trained models to improve annotation robustness. We examine the relationship between sentiment and contextual features such as timestamp, geolocation, and lexical content. To identify latent themes, we apply Latent Dirichlet Allocation (LDA) on partitioned subsets grouped by sentiment and metadata attributes. Finally, we develop an interactive visualization interface to support exploration of sentiment trends and topic distributions across time and regions. This work contributes a scalable methodology for social media analysis in dynamic geopolitical contexts.
\end{abstract}

\section{Introduction}
In recent years, large-scale data annotation has become a critical component in natural language processing (NLP) tasks. However, traditional annotation pipelines often rely on human experts or crowdworkers, which incurs substantial cost and time, especially for high-volume, dynamic domains such as social media. To address this limitation, we adopt a weak supervision approach that leverages a diverse set of well-trained labeling functions. These functions generate noisy but informative labels, which we aggregate using a majority voting scheme. This strategy allows for scalable and low-cost labeling, enabling near real-time analysis—an essential capability for rapidly evolving events like the Russia-Ukraine War. Such timely insights can inform both public understanding and policy responses.

Given that the majority of Twitter users post in English and are influenced by mainstream media narratives, we anticipate an inherent class imbalance in the collected dataset. This imbalance poses a risk to supervised learning frameworks, which are known to be sensitive to skewed label distributions. Instead of relying on manually labeled data or training a new model from scratch, we utilize existing Transformer-based models, which have been shown to capture rich semantic representations even without task-specific fine-tuning. By combining their outputs via majority voting, we further mitigate uncertainty in the labeling process \cite{DBLP:journals/corr/abs-1711-10160}.

Once the dataset is labeled, we examine the relationship between sentiment and various metadata features such as posting date, geolocation, and lexical indicators. To uncover thematic structures within different segments of the data, we apply Latent Dirichlet Allocation (LDA) on subsets grouped by sentiment or other contextual attributes. To support interactive exploration, we also develop a visualization interface that allows users to analyze sentiment distributions, topic trends, and geospatial patterns across time. This work demonstrates a fully automated, cost-efficient pipeline for extracting insights from social media data in dynamic geopolitical contexts.

\section{Related Work}
Representation, as the basis of learning, is very important in sentiment analysis. A good representation form should be efficient, easy to learn and containing enough information. Generally speaking, this task can be done at three levels: document-level, sentence-level and aspect-level, the granularity of which is increasing. For document-level analysis, the Bag of Word model (BoW) is a widely used representation in natural language processing and text mining, which maps each document into a vector of certain length. This vector of document has elements representing the appearance of various words. However, BoW is limited since it ignores the order of words as well as semantic information. To overcome this problem, word embedding technology, which replaces words into low-dimension vectors incorporating syntactic and semantic information, was proposed. Given enough training samples, this representation has been proven to be more powerful and promising.

For sentence-level analysis, sentence embedding, which combines word embeddings and other context information within the sentence, becomes possible due to constraints on the maximum of input length. At the very beginning, handcraft features such as parse tree, opinion lexicon and label of part of speech (pos) were introduced to help model understand a sentence. Nowadays, there are increasing end-to-end models (various types of CNNs and RNNs) able to generate sentence embeddings. For aspect-level, its representation is similar to sentence-level ones. However, the challenge is to determine the dependency between target word and other words in context where positive opinions and negative ones might both exist.

As for the learning part, we mainly focus on aspect-level analysis. As one of the most important factors, target entity information was integrated into sentence embeddings by adaptive recursive neural network (AdaRNN)\cite{dong-etal-2014-adaptive}. On top of that, the semantic relatedness of a target word with its context words was effectively modelled in Target-connection LSTM (TC-LSTM) \cite{2015Effective}, which adopts two LSTM models to process before-target contexts and after-target contexts respectively. Interactive attention network (IAN) \cite{inproceedings}, on the other hand, emphasized the importance of context words and interactively learned attentions in contexts and targets. Graph convolutional network was introduced into this task by ASGCN \cite{zhang-etal-2019-aspect} which combines GCN and dependency tree of sentences. Furthermore, graph ensemble learning framework  was invented to leverage syntactic information over multiple dependency trees in the face of unavoidable errors.

\begin{table}[!t]
\centering
\begin{tabular}{|l|p{11cm}|}
\hline
\textbf{Label} & \textbf{Example} \\
\hline
POS & Johnson and Biden getting so involved in what Russia do is probably more as it is win win for them. Putin invades and they can claim how they told us so and they are a united force against the baddie, Putin doesn't invade and it is 'look we helped we showed a strong stance'. \\
\hline
NEU & Russia: We will not invade Ukraine unless we are provoked. \\
\hline
NEG & David Russell, pointed out that last week, US officials warned that Russia could invade Ukraine at any time throwing the market into chaos, leading to a sell-off in stocks and a rise in oil prices, which relatively helped gold rise above the resistance level of \$1,855 to \$1,860. \\
\hline
\end{tabular}
\caption{Example tweets categorized by sentiment label.}
\label{tab:my-table}
\end{table}

\section{Dataset}

The original raw dataset\footnote{https://www.kaggle.com/datasets/foklacu/ukraine-war-tweets-dataset-65-days} comes from kaggle website. This dataset use twitter api to fetch tweets containing keywords like "ukraine war", "ukraine troops", "ukraine border", "ukraine NATO", "StandwithUkraine", "russian troops", "russian border ukraine", "russia invade" from Jan 1st of 2022 to Mar 6th 2022.

Then we apply a pretrained RoBERTa model\cite{nguyen2020bertweet} which is trained on English tweets to perform sentimental analysis and yields 1.6\% positive tweets, 45.48\% neutral tweets and 52.92\% negative tweets. And we divide the original dataset into three parts with different sentiment. The followed Table \ref{tab:my-table} shows exactly what data in each corpus looks like.

Also, we have extracted key features, like date, location and content, from raw dataset. To know how public opinions variate across the nations, we replace the location with countries by searching with geocoding location API. Finally we get a toll of 1,316,005 tweets.

\section{LDA Analysis}

\subsection{Introduction}

\hspace{1em} Latent Dirichlet allocation (LDA) is one of the most popular topic modeling methods. Suppose there are $D$ Tweets corresponding to a V-word vocabulary. Each tweet consists of $N$ words (removing and padding are allowed). Assume there exist $K$ topics, then the generative procedures of LDA are:

\begin{enumerate}
    \item For each topic $1, \ldots, K$:
\\Draw a multinomial over words $\varphi \sim \operatorname{Dir}(\beta)$
    \item For each Tweet $1, \ldots, D$:
\\Draw a multinomial over topics $\theta \sim \operatorname{Dir}(\alpha)$
\\For each word $w_{N_{d}}$:
 \\\hspace*{1cm}Draw a topic $Z_{N_{d}} \sim \operatorname{Mult}\left(\theta_{D}\right)$ with $Z_{N_{d}} \in[1 \ldots K]$
\\\hspace*{1cm}Draw a word $W_{N_{d}} \sim \operatorname{Mult}(\varphi)$
\end{enumerate}

\noindent
The inference problem in LDA is to compute the posterior of the hidden variables given the Tweet and corpus parameter $\alpha$ and $\beta$.

\begin{figure*}[htbp]
	\centering
	\captionsetup{justification=centering,margin=1cm}
	\includegraphics[width=0.9\linewidth,scale=1]{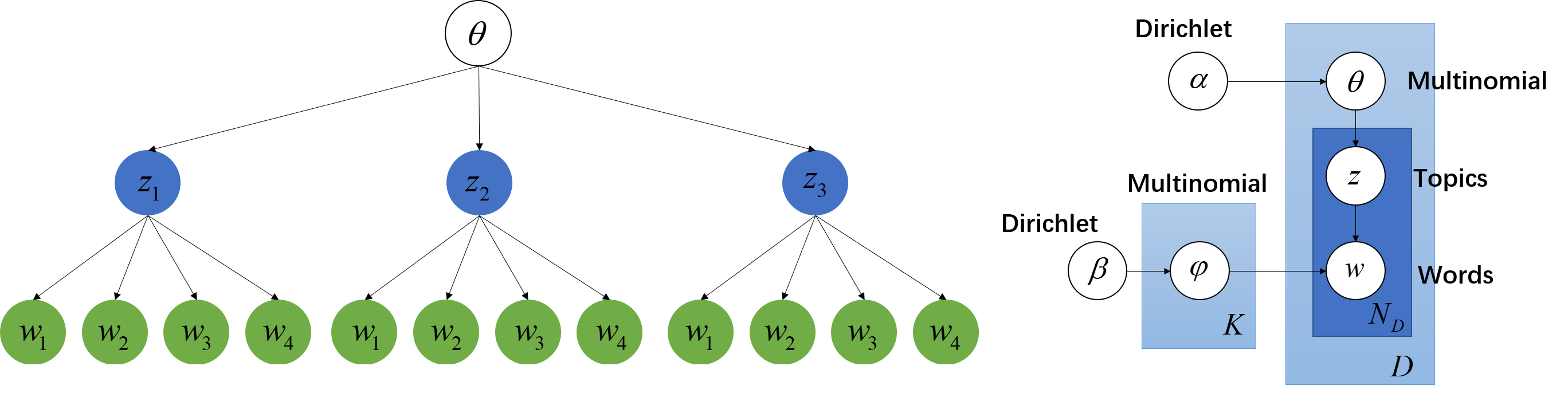}
	\caption{Structure of LDA model}
	\label{Fig0_xsh}
\end{figure*}

\subsection{Result \& Analysis}
\hspace{1em} By using LDA model, we extracted keywords from different aspects. To show the results, we visualized them in WordCloud with the size of words corresponding with their probability in different topics. The first aspect is sentiment: we first counted the number of Tweets related to the three types of sentiments. It can be seen that very few Tweets are positive. Almost half of them are neutral. And the rest is negative. For positive Tweets, people mainly said they would stand with Ukraine. And they express thanks to nations and people who support Ukraine. They also expressed their love for peace. For neutral Tweets, people talked about the reason of the conflict that is Ukraine wants to join Nato. For negative opinions, people mentioned Russian troop. Some of them also criticized Biden's policy such as encourage Ukraine to join Nato. Some of them asked Biden not to buy Russian oil and send troop to help Ukraine.

\begin{figure*}[htbp]
    \centering
    \begin{minipage}[t]{0.3\textwidth}
	\centering
	\includegraphics[width=1\textwidth]{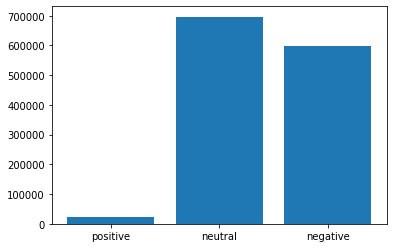}
	\end{minipage}
	\hspace{0.1in}
	\begin{minipage}[t]{0.3\textwidth}
	\centering
	\includegraphics[width=1\textwidth]{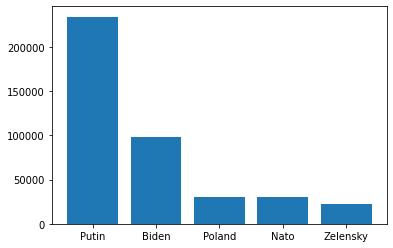}
	\end{minipage}
	\hspace{0.1in}
	 \begin{minipage}[t]{0.3\textwidth}
	\centering
	\includegraphics[width=1\textwidth]{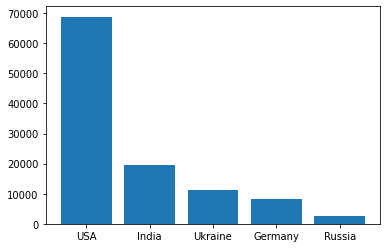}
	\end{minipage}
	\caption{Statistics of related Tweets from different aspects}
	\label{Fig1_xsh}
\end{figure*}

The second aspect is certain topic: we chose several keywords and tried to find the key information for Tweets that include these keywords. Here we give five keywords and counter the Tweets including these keywords. It can be seen that the keyword 'Putin' has the most related Tweets, which is followed by 'Biden' and 'Nato'. For the keyword 'Biden', people mentioned more about his altitude towards Russia and China. Some of them also asked Biden to prevent Russia from invading Ukraine. Quite a few people compared Biden with Trump. For the keyword 'Zelensky', people asked him to save Ukraine people and to recover eastern borders of Ukraine. For the keyword 'Poland', people mentioned more about Nato since Poland is the nearest Nato member to Ukraine. People also asked local governments to help the international students cross the border of Poland.

\begin{figure*}[htbp]
    \centering
	\includegraphics[width=\textwidth]{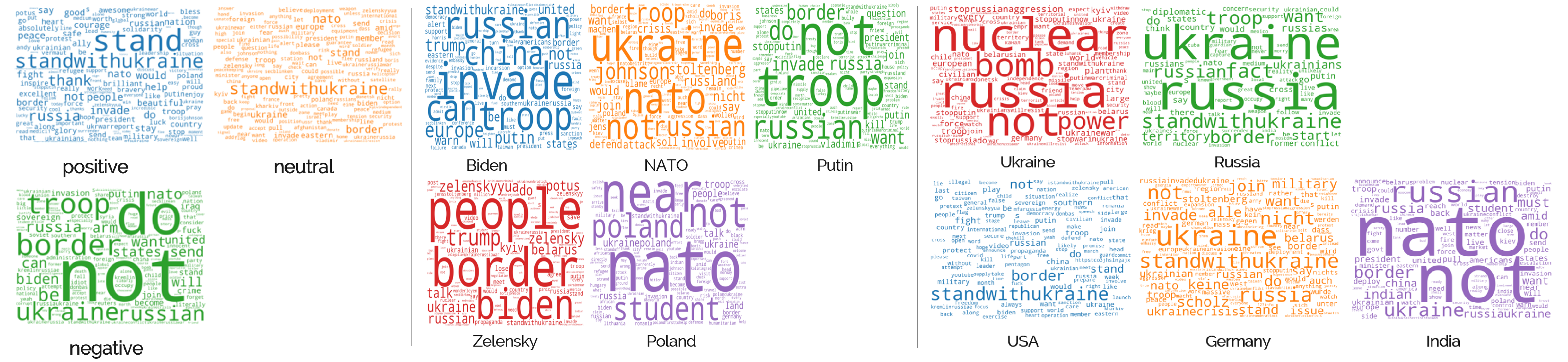}
	\caption{Extracted keywords from different aspects}
	\label{Fig2_xsh}
\end{figure*}

The third aspect is date: we first drew the curve of the number of comments varying with dates. The start time of our dataset is Dec. 31, 2021, while the end time is Mar. 5, 2022. For the special time points on the curve such as peaks and sudden-rising moments, we found related big news on these dates. For example, when NATO put troops on standby on Jan. 24, 2022, the number of related Tweets reached a peak. The keywords extracted on that day show that people talked about the rising tension. And they said Nato had prepared to deploy troops on the border of Poland. Then the number of realted Tweets began to drop until Putin accused West of trying to draw Russia into war on Feb. 2, 2022. People mentioned Putin mostly on that moment. They said Putin's concern about Nato being too close to Russia, and expressed the fear of nuclear war. They also focused on the eastern border of Ukraine. Then when Russia and Belarus began 10 days of military maneuvers, the number of related Tweets began to rise again. And finally it reaches the highest value when Russia started the war on Feb. 24, 2022. On that day, people mention Trump's opinion on Putin and the Russia-Ukraine conflict. They also used the tag "nowar". Some of they considered this event as an invasion and said they would stand with Ukraine.

The last aspect is user location: we chose some representative countries to extracted key information from their users. It can be seen that Ukraine people feared about nuclear power and bombs. They also cared about their children. They hoped someone can stop Putin. For Russian people, some of them also said they would stand with Ukraine. Some of them talked about the occupied territory of Ukraine. Some of them also mention the facts that US is actually involved in this event. They also said Russia did not want invade any country. For American users, they expressed their support for Ukraine. Some of them volunteered to fight with Russia force. Some of them mentioned country leaders like Joe Biden and Zelensky. Some of them also mentioned Trump. For Germans, they also said standing with Ukraine. They also mentioned some European leaders. For Indian people, they talked more about Nato. They also cared about Indian students in Ukraine. 

\begin{figure*}[htbp]
    \centering
	\includegraphics[width=\textwidth]{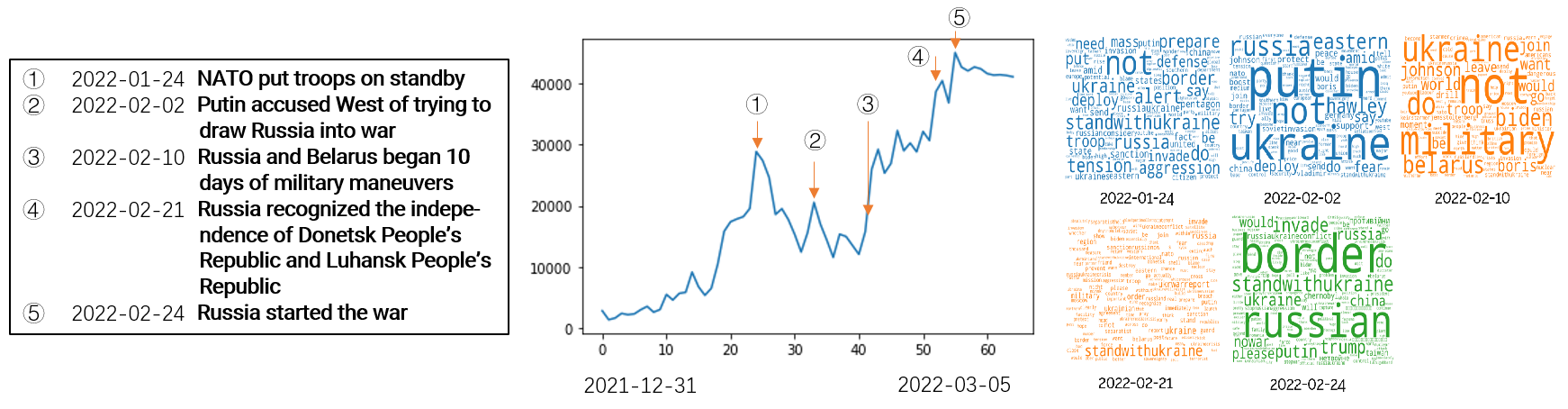}
	\caption{Keyword analysis on different dates}
	\label{Fig3_xsh}
\end{figure*}

To conclude, LDA model is adopted to extract key information from different aspects. By doing this, we are able to find various-dimensional key information in a large text dataset.

\section{Visualization}

We design a UI system to visualize how public views variate across the nations, how it changes as the war still going on, and understand what people really care about. 

\begin{figure*}[htbp]
\begin{minipage}[t]{0.5\textwidth}
	\centering
	\includegraphics[width=0.6\linewidth,scale=1.00]{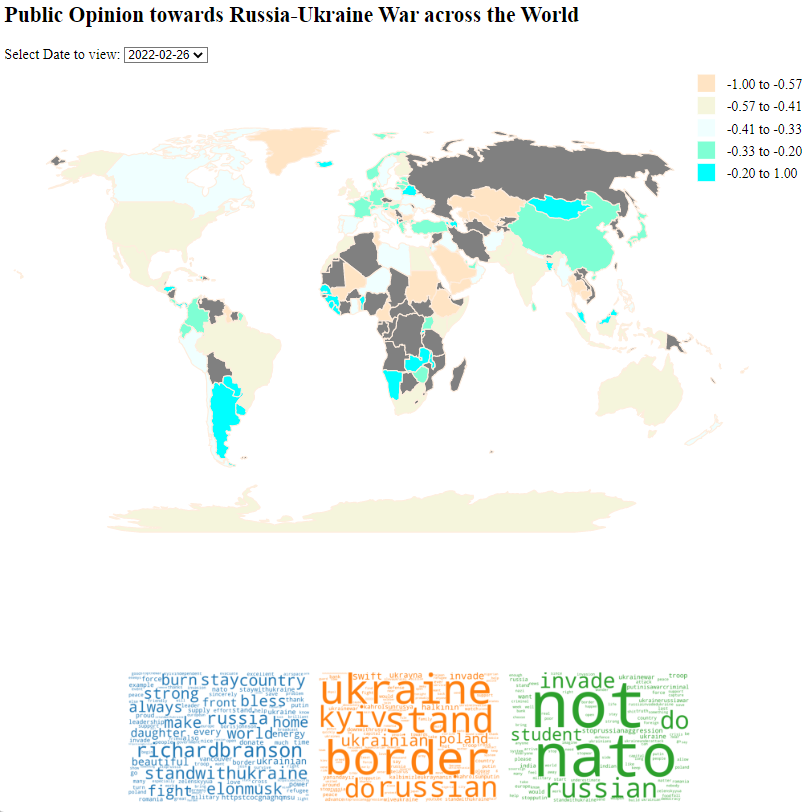}
	\caption{Visualization panel}
	\label{panel}
\end{minipage}
\hspace{0.01in}
\begin{minipage}[t]{0.5\textwidth}
	\centering
	\includegraphics[width=0.95\linewidth,scale=1.00]{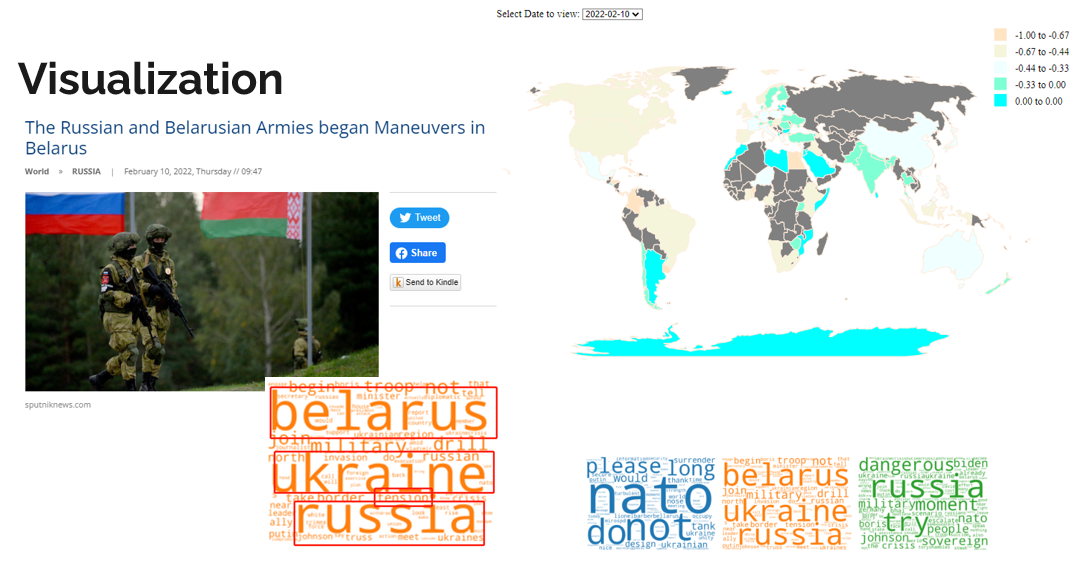}
	\caption{Russia starts a maneuver with Belarus}
	\label{manuever}
\end{minipage}
\end{figure*}

We use the D3 library to write our visualization panel. As you can see in the figure \ref{panel}, there is a world map showing how public opinion variate across different nations. The more blue the color is, the more positive the public opinion people have in the district. Meanwhile, the more white the color is, the more pessimistic the public opinion toward the war. Also, the color grey means at that specific date, we didn't fetch any data from that district, which is acceptable because in some countries people don't share their opinions through Twitter.

From the up-left corner the user can choose a different date to see how the attitude across the nations changed by date. And as the user chooses a different date, keywords that output from the LDA model extracted from the three corpus are shown at the bottom. And from this panel we can see how our model reflects the daily conerns.

\begin{figure*}[htbp]
\begin{minipage}[t]{0.5\textwidth}
	\centering
	\includegraphics[width=0.95\linewidth,scale=1.00]{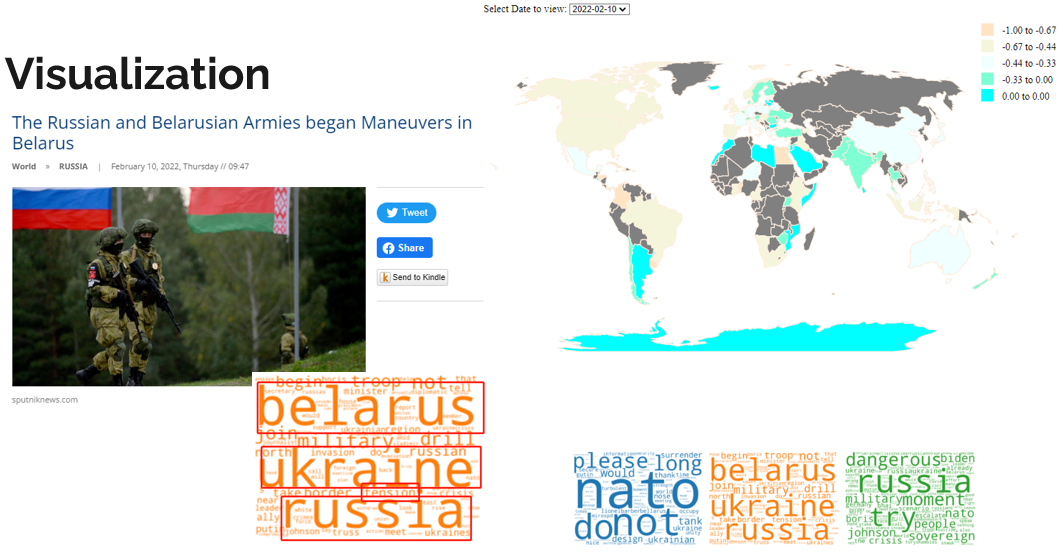}
	\caption{Boris Johnson says Ukraine crisis has entered 'most dangerous moment'}
	\label{boris}
\end{minipage}
\hspace{0.01in}
\begin{minipage}[t]{0.5\textwidth}
	\centering
	\includegraphics[width=0.95\linewidth,scale=1.00]{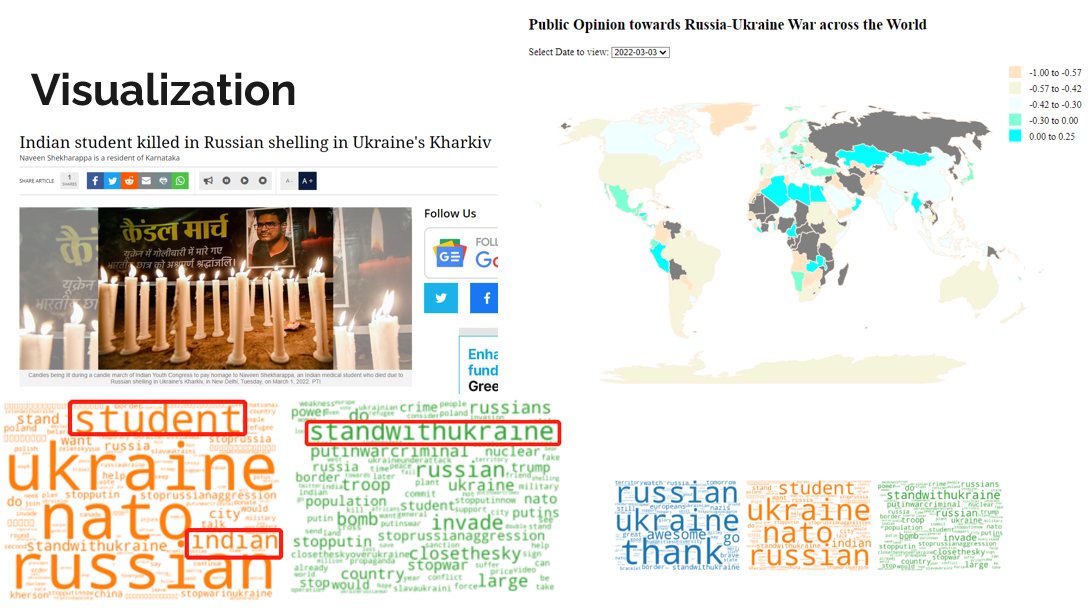}
	\caption{India student was killed by hate crime}
	\label{Indiastudent}
\end{minipage}
\end{figure*}

On the date of Feb 10th, The Russia starts a maneuver with Belarus. As we can see from Figure \ref{manuever}, three keywords about the news, which is "belarus","Ukraine" and "Russia" is catched by our LDA model with neutral corpus. Figure \ref{boris} shows on the same day there is another pessimistic news with title "Boris Johnson says Ukraine crisis has entered 'most dangerous moment'", and keywords, "Boris Johnson", "dangerous" and "moment" are reflected in our LDA output with negative corpus input. 

Our LDA model can not only extract some campaign slogans like "standwithUkraine", but can also extract some news draw extensively concerns on social media. As we can see in Figure \ref{Indiastudent}, on Mar 3rd, just several days after Putin launched the war, hashtag like "standwithUkraine" was widely detected on the corpus and output by our LDA model. At the same day another tragedy happened that an Indian student was killed because of hate crime and keywords like "Indian" and "student" are also extracted by our LDA model.

\section{Discussion}

In our project, LDA does help for public opinion extraction. We successfully see (1) How public opinions variate from different district (2) what people think about different topics (3) what tweets with different emotions care about (4) How the focus of public opinion changes over time.

However, there exists some limitations in our project. First of all, our data is fetched from twitter, which means we could neglect public opinions from non-English speaking world. For example, twitter is blocked in Russia by local government, so actually we can not derive many tweets from Russian users. To mitigate such problem, more data needs to be fetched from other media platform. 

Also, there may exist many bots who just keep sending tweets under some specific hashtags to influence public opinions. So actually we can not decide exactly whether a tweet is sent by human beings or bots. To mitigate this issue, we can just delete same tweets send by the same user.

In addition, there are some limitations for LDA model. Due to the linearity of the technique, LDA is not scalable. It also ignores topic evolving in the Tweets, as well as co-occurrence relations between different Tweets, for example, when someone retweets other people's opinions. It may fail to deal with data with high information entropy since it requires distinct word distributions under different topics.

\bibliographystyle{acl_natbib}
\bibliography{anthology,acl2021}

\end{document}